\documentclass[conference,a4paper]{IEEEtran}
\IEEEoverridecommandlockouts

\usepackage[dvips]{graphicx}
\usepackage{amsmath,amssymb}
\usepackage{paralist}
\usepackage{eclbkbox}
\usepackage{subfigure}
\usepackage{url}
\usepackage{fancyhdr}

\makeatletter
\def\tbcaption{\def\@captype{table}\caption}
\def\figcaption{\def\@captype{figure}\caption}

\newcommand{\bvec}[1]{\mbox{\boldmath $#1$}}

\begin{document}
% paper title
\title{Fine Tuning Method by using Knowledge Acquisition from Deep Belief Network
\thanks{\copyright 2016 IEEE. Personal use of this material is permitted. Permission from IEEE must be obtained for all other uses, in any current or future media, including reprinting/republishing this material for advertising or promotional purposes, creating new collective works, for resale or redistribution to servers or lists, or reuse of any copyrighted component of this work in other works.}
}

\author{
\IEEEauthorblockN{Shin Kamada}
\IEEEauthorblockA{Dept. of Intelligent Systems\\
Graduate School of Information Sciences\\
Hiroshima City University\\
3-4-1, Ozuka-Higashi, Asa-Minami-ku\\
Hiroshima, 731-3194, Japan\\
E-mail: da65002@e.hiroshima-cu.ac.jp}
\and
\IEEEauthorblockN{Takumi Ichimura}
\IEEEauthorblockA{Dept. of Management and Systems,\\
Prefectural University of Hiroshima\\
\\
1-1-71, Ujina-Higashi, Minami-ku, \\
Hiroshima 734-8558, Japan\\
E-mail: ichimura@pu-hiroshima.ac.jp}
}

\maketitle

\pagestyle{fancy}{
\fancyhf{}
\fancyfoot[R]{}}
\renewcommand{\headrulewidth}{0pt}
\renewcommand{\footrulewidth}{0pt}

\begin{abstract}
We developed an adaptive structure learning method of Restricted Boltzmann Machine (RBM) which can generate/annihilate neurons by self-organizing learning method according to input patterns. Moreover, the adaptive Deep Belief Network (DBN) in the assemble process of pre-trained RBM layer was developed. The proposed method presents to score a great success to the training data set for big data benchmark test such as CIFAR-10. However, the classification capability of the test data set, which are included unknown patterns, is high, but does not lead perfect correct solution. We investigated the wrong specified data and then some characteristic patterns were found. In this paper, the knowledge related to the patterns is embedded into the classification algorithm of trained DBN. As a result, the classification capability can achieve a great success (97.1\% to unknown data set).
\end{abstract}


\IEEEpeerreviewmaketitle

\section{Introduction}
Deep Learning is well known to be the representative method of artificial intelligence. The representation learning can discover the good set of features to input patterns and calculate the representation itself. Many kinds of structures and learning methods have been developed to achieve the great success. It is often said that Deep Learning should include the hierarchical model deeply, or the discovering of optimal structure and its parameters of Convolutional Neural Network (CNN) \cite{LeCun89} is important. This issue pointed out by many researchers is right definitely, however, the effort to find the optimal structure and the parameters is very expensive and the calculation cost becomes high. To realize high level representation at low calculation cost, the self-organizing mechanism to adjust the structure itself and parameters simultaneously should be required with the statistical learning method.

We have developed the structural learning method of Restricted Boltzmann Machine (RBM) \cite{Hinton12} by neuron generation/annihilation algorithm \cite{Kamada16_ICONIP}. The adaptive learning method that can discover the optimal number of hidden neurons according to the input space is important method in terms of the stability of energy function. In \cite{Kamada16_ICONIP}, Structural Learning Method with Forgetting (SLF) proposed by Ishikawa \cite{Ishikawa96} was embedded to make clear interpretation, because the regularization is used to induce models to be sparse and to prevent over fitting. 

Deep Belief Network (DBN) \cite{Hinton06} has a deep architecture that can represent multiple features of input patterns hierarchically with the pre-trained RBM. It was natural that we reached the development of the adaptive learning method of DBN \cite{Kamada16_TENCON}. The method can determine an optimal number of hidden layers during the learning. The learning method of DBN showed the high classification capability of benchmark test CIFAR-10 and CIFAR-100 \cite{CIFAR10}. Especially, the result on CIFAR-10 showed almost 100.0\% to the training data set, but 92.4\% to the test data set. The rate for the test data set is not the top score in \cite{CIFAR10_result}. We investigated the structure and the parameters of the computation result to discover the misclassification reason \cite{Kamada16_YRW}. As a result, the ambiguous images such as `dog' and `cat' cause the misclassification result. The path ways in DBN given data of misclassification are traced and the nodes which the 2 paths diverge are found. The fire rule of dividing node is defined and is embedded to the trained DBN. As a result, the classification capability can achieve a great success (97.1\% to unknown data set on CIFAR-10, this is the top score of world records in \cite{CIFAR10_result}).

The integration of knowledge into the trained deep neural network shows the effectiveness in the research field of natural language processing (NLP) \cite{Bollegala16}. NLP is required large text corpora to learn the word representations. If the frequency of corpora is low, the accuracy of the model does not improve. On the other hand, the traditional model using the knowledge database cannot respond to the question, if the word in the question is involved in the database. Many kinds of fusion model of neural language model and semantic model has been challenged to realize human like interpretation.

We happen to meet the misclassification to the ambiguous image data, but our challenge brings forth a new approach that gives the reinforcement stimulus to the trained network due to the flow of composed signal in the neural model. As a result, we can achieve success to the best performance to unknown data \cite{CIFAR10_result}.

\section{Adaptive Structure Learning of DBN}
\subsection{Overview of RBM}
RBM is energy-based statistical model for unsupervised learning. It has the network structure with 2 kinds of layers where one is a visible layer $\bvec{v} \in \{ 0, 1 \}^{I}$ for input data and the other is a hidden layer $\bvec{h} \in \{ 0, 1 \}^{J}$ for representing the features of given data space. There are the connections between neurons except the neurons in same layer. The RBM learning employs to train the some parameters $\bvec{\theta}=\{\bvec{b}, \bvec{c}, \bvec{W} \}$ till the energy function $E(\bvec{v}, \bvec{h})$ becomes to a certain small value. $p(\bvec{v}, \bvec{h})$ is the joint probability distribution of $\bvec{v}$ and $\bvec{h}$.
\begin{equation}
E(\bvec{v}, \bvec{h}) = - \sum_{i} b_i v_i - \sum_j c_j h_j - \sum_{i} \sum_{j} v_i W_{ij} h_j ,
\label{eq:energy}
\end{equation}

\begin{equation}
p(\bvec{v}, \bvec{h})=\frac{1}{Z} \exp(-E(\bvec{v}, \bvec{h})), \;  Z = \sum_{\bvec{v}} \sum_{\bvec{h}} \exp(-E(\bvec{v}, \bvec{h})) ,
\label{eq:prob}
\end{equation}
where $b_i$ and $c_j$ are the parameters for $v_i$ and $h_j$, respectively. $W_{ij}$ is the weight between $v_i$ and $h_j$. $Z$ is the partition function which is given by summing over all possible pairs of visible and hidden vectors. The optimal parameters $\bvec{\theta}=\{\bvec{b}, \bvec{c}, \bvec{W} \}$ according to the distribution of given input data can be calculated by partial derivative of $p(\bvec{v})$ (that is maximum likelihood estimation). Contrastive Divergence (CD) \cite{Hinton02} is well known RBM learning method to be a faster algorithm of Gibbs sampling based on Markov chain Monte Carlo methods. The convergence situation of RBM learing with CD binary sampling was discussed under the Lipschitz continuous by Carlson et al \cite{Carlson15}. Moreover, we investigated the change of gradients for the parameters $\bvec{\theta}=\{\bvec{b}, \bvec{c}, \bvec{W} \}$ during the learning phase, then the 2 kinds of parameters $\bvec{c}$ and $\bvec{W}$ are influenced on the convergence situation of RBM except the parameter $\bvec{b}$ because the gradient for parameter $\bvec{b}$ may be affected by various features of input patterns even if it includes an noise \cite{Kamada16-SMC}.

\subsection{Neuron Generation and Annihilation Algorithm}
\label{subsec:AdaptiveRBM}
\begin{figure}[tbp]
\begin{center}
\subfigure[Neuron Generation]{\includegraphics[scale=0.5]{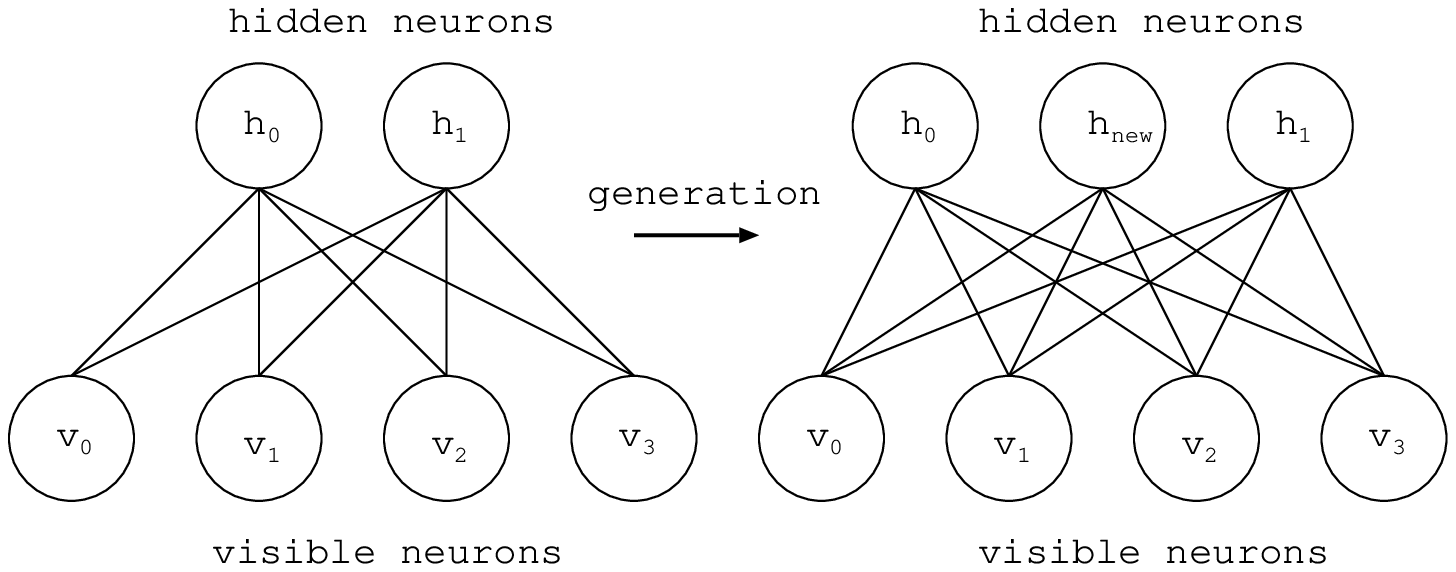}
  \label{fig:neuron_generation}
}
\subfigure[Neuron Annihilation]{\includegraphics[scale=0.5]{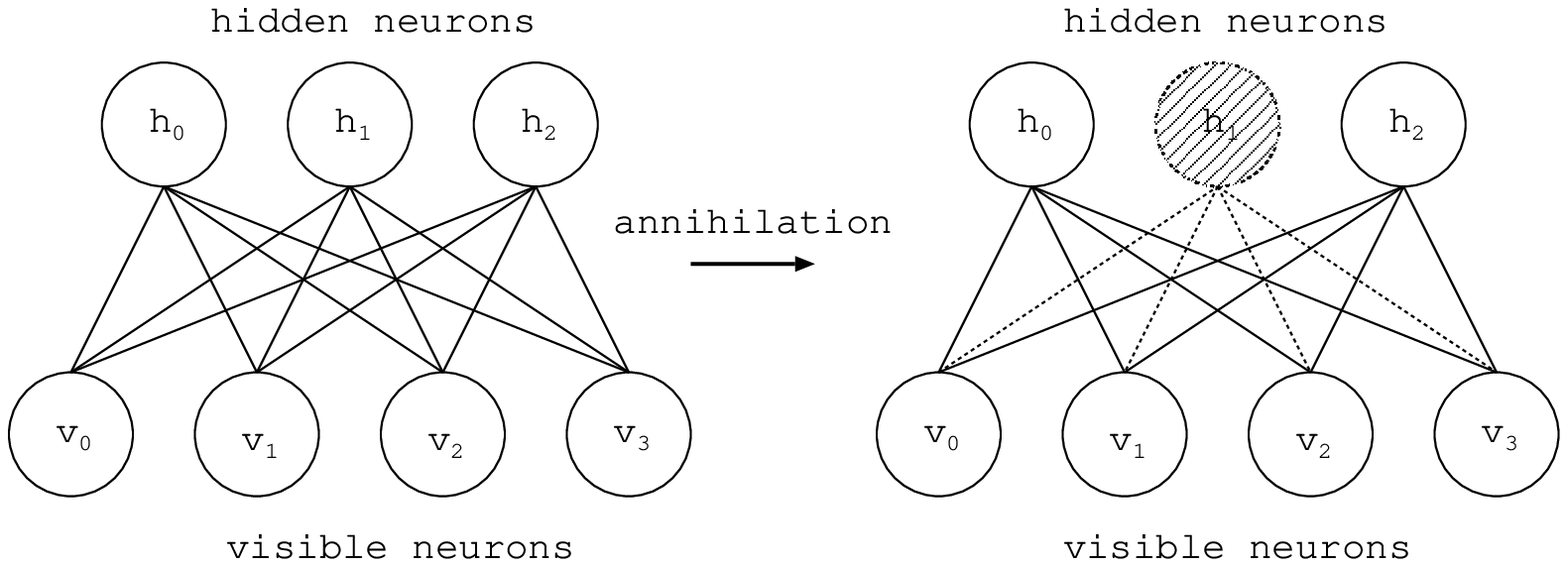}
  \label{fig:neuron_annihilation}
}
\caption{Adaptive RBM}
\label{fig:adaptive_rbm}
\end{center}
\end{figure}

We have proposed the adaptive learning method of RBM that the optimal number of hidden neurons can be self-organized according to the features of a given input data set in learning phase. The basic idea of the neuron generation and annihilation algorithm of RBM is inspired by \cite{Ichimura04} in multi-layered neural network, that is to measure the network stability with the fluctuation of weight vector called the Walking Distance ($WD$) during the learning phase. If a neural network does not have enough neurons to be satisfied to infer, then $WD$ will tend to fluctuate greatly even after a certain period of the training process, because some hidden neurons may not represent an ambiguous pattern due to lack of the number of hidden neurons. In such a case, we can solve the problem by dividing a neuron which tries to represent the ambiguous patterns into 2 neurons by inheriting the attributes of the parent hidden neuron. Then we use the condition of neuron generation with inner product of the variance of monitoring 2 parameters $\bvec{c}$ and $\bvec{W}$ except $\bvec{b}$ as shown in Eq.(\ref{eq:neuron_generation}).
\begin{equation}
(\alpha_{c} \cdot dc_j) \cdot (\alpha_{W} \cdot dW_{ij} )> \theta_{G},
\label{eq:neuron_generation}
\end{equation}
where $dc_j$ and $dW_{ij}$ are the gradient vectors of the hidden neuron $j$ and the weight vector between neuron $i$ and $j$, respectively. $\alpha_{c}$ and $\alpha_{W}$ are the constant values for the adjustment of the range of each parameter. $\theta_{G}$ is an appropriate threshold value. A new hidden neuron will be generated and inserted into the neighborhood of the parent neuron as shown in Fig.~\ref{fig:neuron_generation} if Eq.(\ref{eq:neuron_generation}) is satisfied.

After the neuron generation process, the network with some inactivated neurons that don't contribute the classification capability should be removed in terms of the reduction of calculation. If Eq.(\ref{eq:neuron_annihilation}) is satisfied in learning phase, the corresponding neuron will be annihilated as shown in Fig.~\ref{fig:neuron_annihilation}. 
\begin{equation}
\frac{1}{N}\sum_{n=1}^{N} p(h_j = 1 | \bvec{v}_n) < \theta_{A},
\label{eq:neuron_annihilation}
\end{equation}
where $\bvec{v}_{n}=\{ \bvec{v}_{1},\bvec{v}_{2},\cdots,\bvec{v}_{N}\}$ is a given input data, $N$ is the number of samples of input data. $p(h_j = 1 | \bvec{v}_n)$ means a conditional probability of $h_j \in \{ 0, 1 \}$ under a given $\bvec{v}_n$. $\theta_{A}$ is an appropriate threshold value. Fig.~\ref{fig:neuron_annihilation} shows the structure of neuron annihilation.

\subsection{Structural Learning Method with Forgetting}
Although the optimal number of hidden neurons is determined by the neuron generation and annihilation algorithms, we may meet another difficulty that the trained network is too a black box to extract some explicit knowledge from the trained network. In order to solve the difficulty, we developed the structural learning method with forgetting (SLF) to discover the regularities of the RBM network structure \cite{Kamada16_ICONIP}. The basic concept is derived by Ishikawa \cite{Ishikawa96}, that is 3 kinds of penalty terms are added into an objective function $J$ (the CD sampling cost is usually used for RBM learning) as shown in Eq.(\ref{eq:forgetting1}) -  Eq.(\ref{eq:forgetting3}).

Eq.(\ref{eq:forgetting1}) is to decay the weights norm multiplied by small criterion $\epsilon_{1}$. Eq.(\ref{eq:forgetting2}) is used to extract characteristic patterns from the outputs of network by forcing each hidden unit to be fully 0 or 1. Eq.(\ref{eq:forgetting3}) is used to prevent an objective function to be larger than an original one.

\begin{equation}
  \label{eq:forgetting1}
  J_{f} = J + \epsilon_{1} \| \bvec{W} \|,
\end{equation}
\begin{equation}
  \label{eq:forgetting2}
  J_{h} = J + \epsilon_{2} \sum_{i} \min \{ 1 - h_i, h_i\},
\end{equation}
\begin{eqnarray}
  \label{eq:forgetting3}
  J_{s} = J - \epsilon_{3} \| \bvec{W}^{'} \|, \ 
  W^{'}_{ij} = \left\{
  \begin{array}{ll}
    W_{ij}, & if \ |W_{ij}| < \theta \\
    0, & otherwise
  \end{array}
  \right.,
\end{eqnarray}
where, $\epsilon_{1}$, $\epsilon_{2}$ and $\epsilon_{3}$ are the small criterion for each equation. After an optimal number of hidden neurons is determined by neuron generation and annihilation algorithm, both Eq.(\ref{eq:forgetting1}) and Eq.(\ref{eq:forgetting2}) should be applied simultaneously during a certain learning period. Alternatively, Eq.(\ref{eq:forgetting3}) is used instead of Eq.(\ref{eq:forgetting1}) at final learning period in order to prevent an objective function to be larger.

\subsection{Layer Generation Condition}
\label{subsec:DBN}
Deep Belief Network (DBN) is a major deep learning method proposed by Hinton \cite{Hinton06}. The hierarchical DBN network structure becomes to represent higher and multiple level features of input patterns by building up the pre-trained RBM. Energy $E^{l}$ and the conditional probability of hidden neuron at $l$-th layer can be calculated as the following equations.

\begin{equation}
\label{eq:energy_dbn}
E^{l} = E(\bvec{h}^{l-1}, \bvec{h}^{l}) = - (\bvec{b}^{l})^{T} \bvec{h}^{l-1} - (\bvec{c}^{l})^{T} \bvec{h}^{l} - \bvec{h}^{l-1} \bvec{W}^{l} \bvec{h}^{l} ,
\end{equation}
\begin{equation}
\label{eq:prob_dbn}
p(h_j^{l} = 1 | \bvec{h}^{l-1})= sigm(\bvec{c}^{l} + \bvec{W}^{l} \bvec{h}^{l-1}) ,
\end{equation}
where, $\bvec{b}^{l}$, $\bvec{c}^{l}$, $\bvec{W}^{l}$ are the parameters of visible neurons, hidden neurons, and weight in $l$-th layer. $\bvec{h}^{l} \in \{0, 1\}$ is the hidden neurons in $l$-th layer.

We have proposed the adaptive learning method of DBN that can determine the optimal number of hidden layers \cite{Kamada16_TENCON}. Each RBM employs the adaptive learning method by the neuron generation and annihilation algorithm described in section \ref{subsec:AdaptiveRBM}. In general, data representation of DBN performs the specified features from abstract to concrete at each layer in the direction to output layer. That is, the lower layer has the power of non figurative representation, and the higher layer constructs the object to figure out an image of input patterns. Adaptive DBN can automatically adjust self-organization of structured data representation. In the learning process, we observe the total WD (the variance of both $\bvec{c}$ and $\bvec{W}$, the exception of $\bvec{b}$ is for the control of the variance of input data) and energy function. If overall WD is larger than a threshold value and the energy function is still large, then a new RBM is required to express the suitable network structure for the given input data. That is, a large WD and energy function mean that the RBM is the lack figuring capability for the input data. Therefore, we define the condition of layer generation as both the total WD and the energy function as following equations.
\begin{equation}
\sum_{l=1}^{k} (\alpha_{WD} \cdot WD^{l}) > \theta_{L1},
\label{eq:layer_generation1}
\end{equation}
\begin{equation}
\sum_{l=1}^{k} (\alpha_{E} \cdot E^{l}) > \theta_{L2},
\label{eq:layer_generation2}
\end{equation}
where $WD^{l}$ is the total variance of parameters $\bvec{c}$ and $\bvec{W}$ in $l$-th layer. $E^{l}$ is the total energy function in $l$-th layer. $k$ is the current layer. $\alpha_{WD}$ and $\alpha_{E}$ are the constant values for the adjustment of the repercussion of deviant range of $WD^{l}$ and $E^{l}$. $\theta_{L1}$ and $\theta_{L2}$ are the appropriate threshold values. If Eq.(\ref{eq:layer_generation1}) and Eq.(\ref{eq:layer_generation2}) are satisfied simultaneously in learning phase at layer $k$, a new hidden layer $k+1$ will be generated after the learning at layer $k$ is finished. The initial RBM parameters $\bvec{b}$, $\bvec{c}$ and $\bvec{W}$ at the generated layer $k+1$ are set by inheriting the attributes of the parent(lower) RBM.

\subsection{Experimental Results}
\label{sec:EXE}
Experimental results to CIFAR-10 \cite{CIFAR10} benchmark data set are explained in this section. The data set is the image classification of about 60,000 color images categorized into 10 classes, which consists of 50,000 images for the training set and test set for the remaining. The image data used in the experiment was processed by ZCA whitening process in \cite{Dieleman12}. The workstation with the following specifications was used:  CPU = Intel(R) 24Core Xeon E5-2670 v3 2.3GHz, GPU = Tesla K80 4992 24GB $\times$ 3, Memory = 64GB, OS = Centos 6.7 64 bit.

Table \ref{tab:result-correct-ratio} shows the experimental result by using traditional RBM \cite{Dieleman12}, traditional DBN \cite{Krizhevsky10}, simple CNN \cite{Clevert15}, and our proposed methods. In this paper, we prepared Adaptive RBM, Adaptive RBM with Forgetting, and Adaptive DBNs with 2 kinds of parameters. The explanation of the parameters in this paper is described as follows. The training algorithm is Stochastic Gradient Descent (SGD) method, the batch size is 100, the learning rate is 0.1, and the initial number of hidden neurons is 300, $\theta_{A} = 0.1$, $\theta_{L1} = 0.05$, $\theta_{L2} = 0.05$, $\epsilon_{1} = \epsilon_{2} = \epsilon_{3} = 0.01$, $\theta = 0.1$. Table \ref{tab:result-correct-ratio} is the correct ratio to training data set and test data set, respectively. Of course, the classification accuracy to training data set is almost 100\%, because the network model without error was constructed by the training algorithm. However, the test data set have never seen for the given trained network model. Therefore, the current research themes shift to the classification capability to unknown data.

The experimental result shows that our methods are superior to the traditional methods. The high correct ratio depends on the value of the parameters. If the parameter $\theta_{G}$ is small, the self organizing function works well and then the network grows to adjust to the size and the complexity of data.

Adaptive DBN method with 5 layered RBM was constructed. In the network structure, the 2nd and the 4th RBM layer has more neurons than the 3rd and the 5th layer. We consider that the coarse-grained particle of information is stored in the even layer by pre-training. In the higher layer, the combination of the particles is represented and transformed from the condition of being abstract to the concrete sample of the classes.

\begin{table}[h]
\caption{Classification Accuracy}
\vspace{-5mm}
\label{tab:result-correct-ratio}
\begin{center}
\begin{tabular}{l|r|r}
\hline \hline
& \multicolumn{1}{c|}{Training} & \multicolumn{1}{c}{Test}  \\ \hline\hline
Traditional RBM \cite{Dieleman12} &     -     & 63.0\%  \\ \hline
Traditional DBN \cite{Krizhevsky10}      &     -     & 78.9\%  \\ \hline 
Simple CNN \cite{Clevert15}      &     -     & 96.53\%  \\ \hline \hline
Adaptive RBM                 &   99.9\%  & 81.2\%   \\ \hline
Adaptive RBM with Forgetting  &   99.9\%  & 85.8\%   \\ \hline
Adaptive DBN with Forgetting &  100.0\%  & 92.4\%   \\
(No. layer = 5, $\theta_G = 0.05$) & & \\\hline
Adaptive DBN with Forgetting&  100.0\%  & {\bf 97.1\%}   \\
(No. layer = 5, $\theta_G = 0.01$ ) & & \\\hline
\hline 
\end{tabular}
\end{center}
\end{table}

\section{Knowledge Acquisition from the trained DBN}
\label{sec:Knowledge}
\begin{figure*}[tbp]
\centering
\subfigure[Correct case (`Cat' to `Cat')]{\includegraphics[scale=0.9]{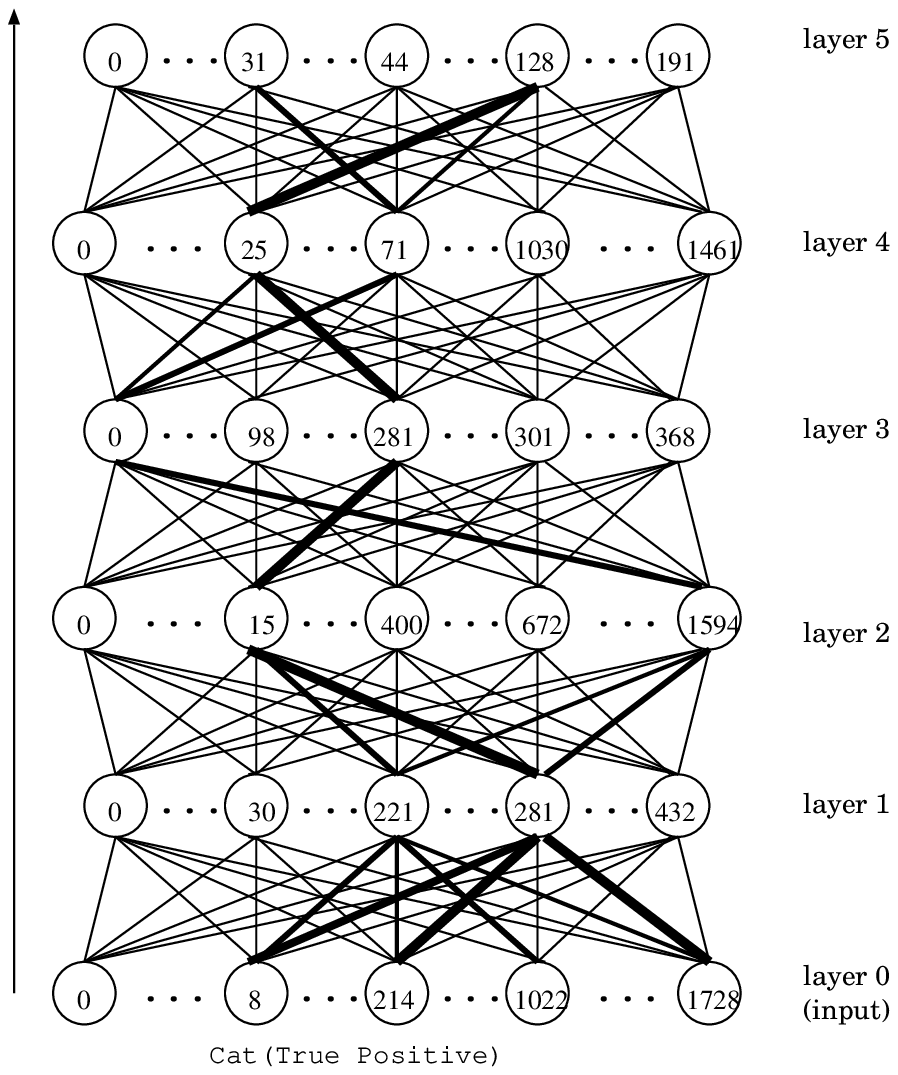}\label{fig:path_3-3}}
\subfigure[Wrong case (`Cat' to `Dog)]{\includegraphics[scale=0.9]{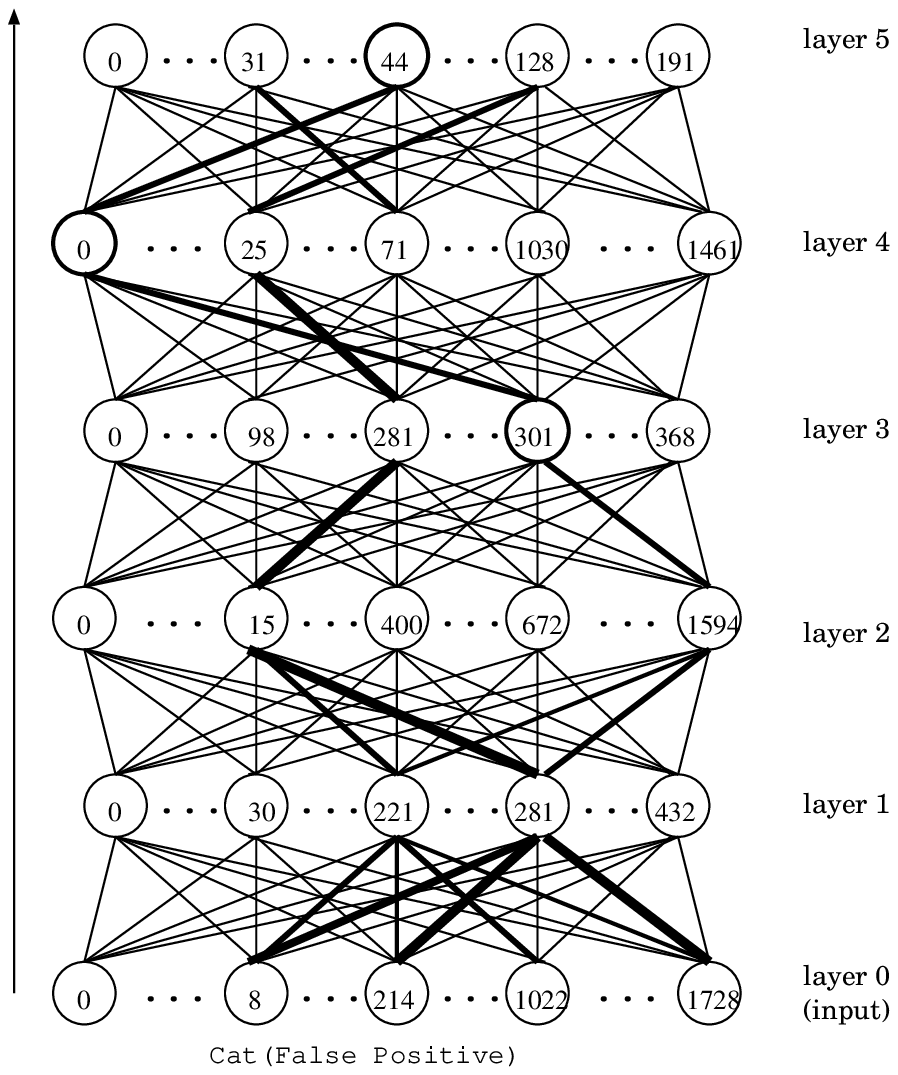}\label{fig:path_3-5}}
\subfigure[Correct case (`Dog' to `Dog')]{\includegraphics[scale=0.9]{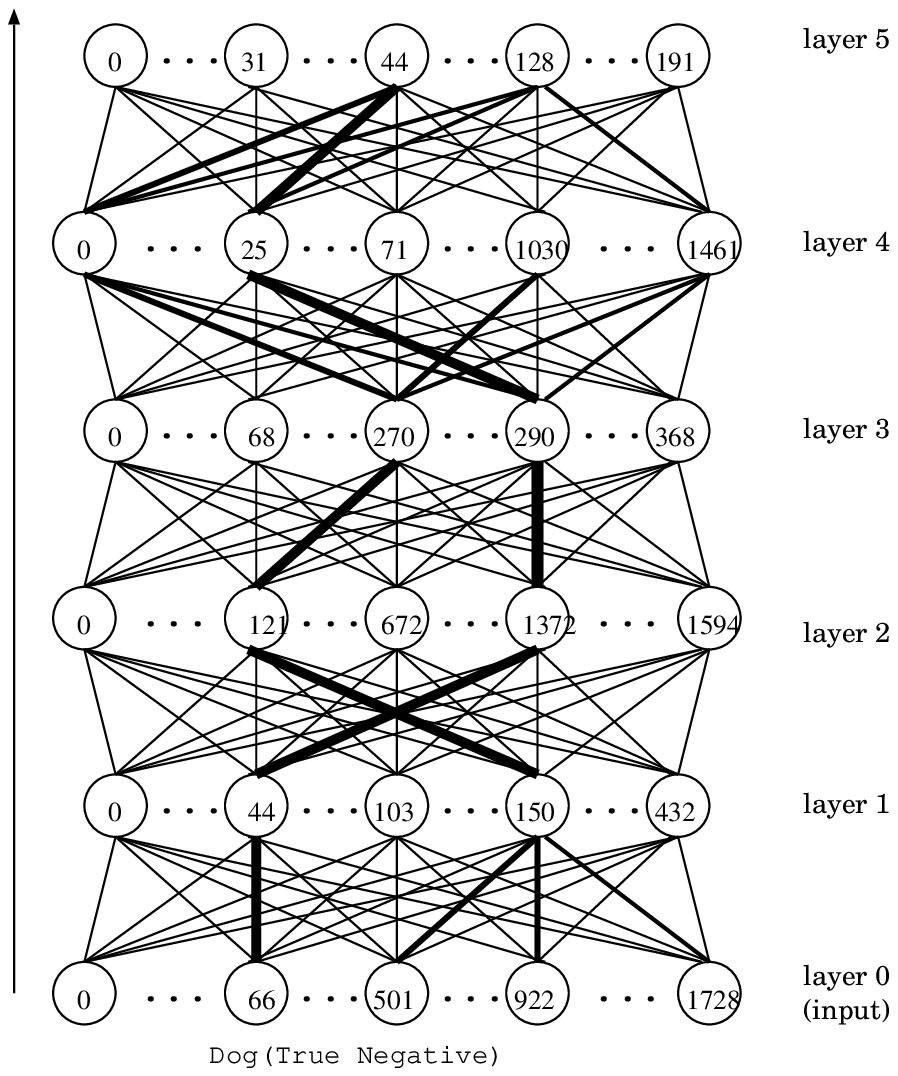}\label{fig:path_5-5}}
\subfigure[Wrong case (`Dog' to `Cat')]{\includegraphics[scale=0.9]{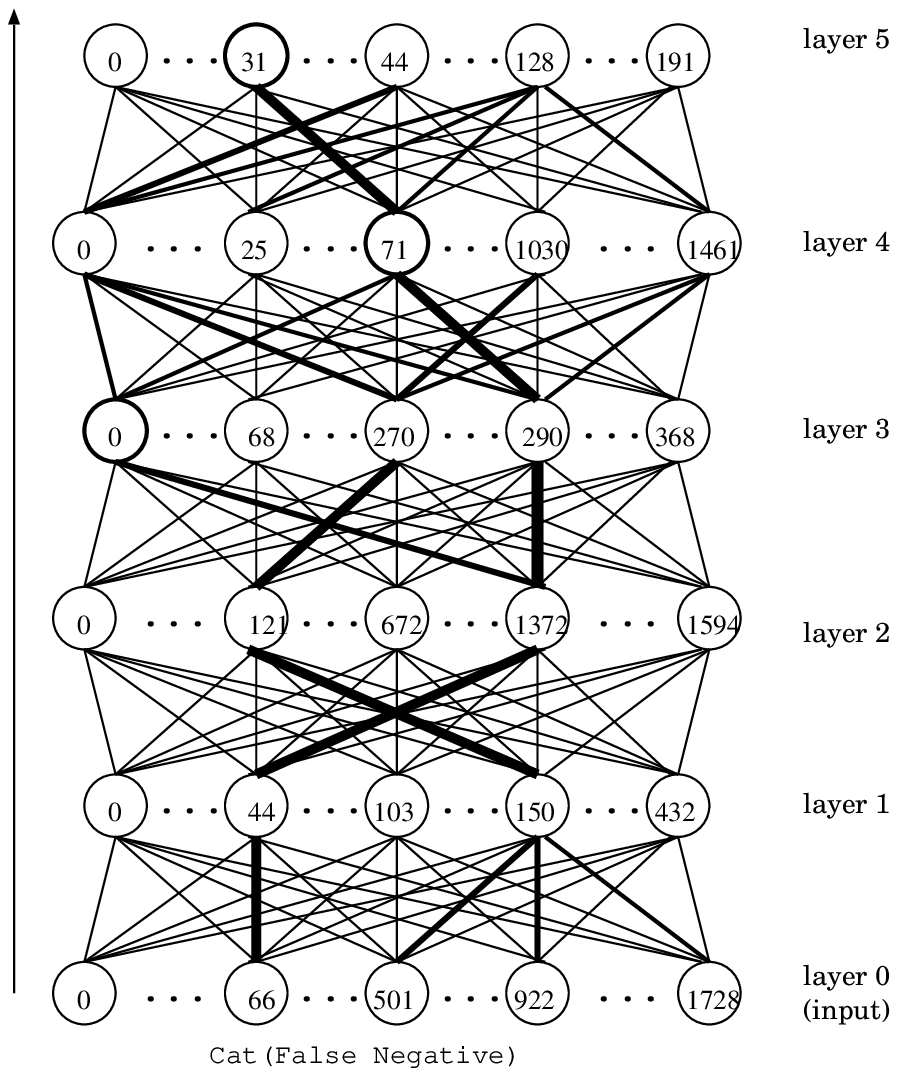}\label{fig:path_5-3}}
\vspace{-3mm}
\caption{The path of activated hidden neurons}
\label{fig:h_act_path}
\end{figure*}

The highest score by Adaptive DBN with Forgetting was 97.1\% and keeps the highest classification capability to unknown data set \cite{CIFAR10_result}(This is the top score of world records as of Aug. 6, 2016). The good result was examined with the varying parameters many times over (about 10 times). However, the discovery of optimal parameter may be difficult without experience or training. Therefore, we investigated the network condition and the images of test set of misclassification. The misclassified images involve ambiguous patterns. For example, same images of `Cat' in the test set are actually classified into `Dog', we will be likely to make a wrong judgment to these images. The misclassification of such cases by deep learning model is natural outcome. In the network model, the misclassification means that the signal goes along the wrong path in the network. Therefore, we traced the path of fired hidden neurons by giving the test data set to the trained DBN and found the paths as shown in Fig.\ref{fig:h_act_path}. 

Fig.\ref{fig:h_act_path} is the visualization of the connection between activated hidden neurons given the test data to the trained DBN. Fig.\ref{fig:path_3-3} and Fig.\ref{fig:path_5-5} are the cases of collect classification for `Cat' and `Dog', respectively. Fig.\ref{fig:path_3-5} and Fig.\ref{fig:path_5-3} are the cases of wrong classification for them, that is `Cat' to `Dog' and `Dog' to `Cat', respectively. The circled node means an hidden neuron, the value in the node means the numbering neuron in each layer. The connection between nodes has 5 stage bold lines representing the strength of weight. 

From the cases of same category as shown in Fig.\ref{fig:path_3-3} (`Cat' to `Cat') and Fig.\ref{fig:path_3-5} (`Cat' to `Dog') or Fig.\ref{fig:path_5-5} (`Dog' to `Dog') and Fig.\ref{fig:path_5-3} (`Dog' to `Cat'), same hidden neurons in lower layers (less than 2 layer) were fired, while neurons dividing into different cases were found in higher layers (more than layer 3). For example, the path from 1594 to 301 between layer 2 and layer 3 was not fired in Fig.\ref{fig:path_3-3}, but the corresponding path was fired in Fig.\ref{fig:path_3-5}. The wrong output pattern connected to the neuron was transferred to the top layer.

Moreover, same output pattern was found between Fig.\ref{fig:path_3-3} (`Cat' to `Cat') and Fig.\ref{fig:path_5-3} (`Dog' to `Cat') or Fig.\ref{fig:path_3-5} (`Cat' to `Dog') and Fig.\ref{fig:path_5-5} (`Dog' to `Dog').

\section{Synthesis of the network model and the acquired rules}
\label{sec:synthesis}
\begin{figure}[tbp]
\centering
\subfigure[Fired neurons]{\includegraphics[scale=0.22]{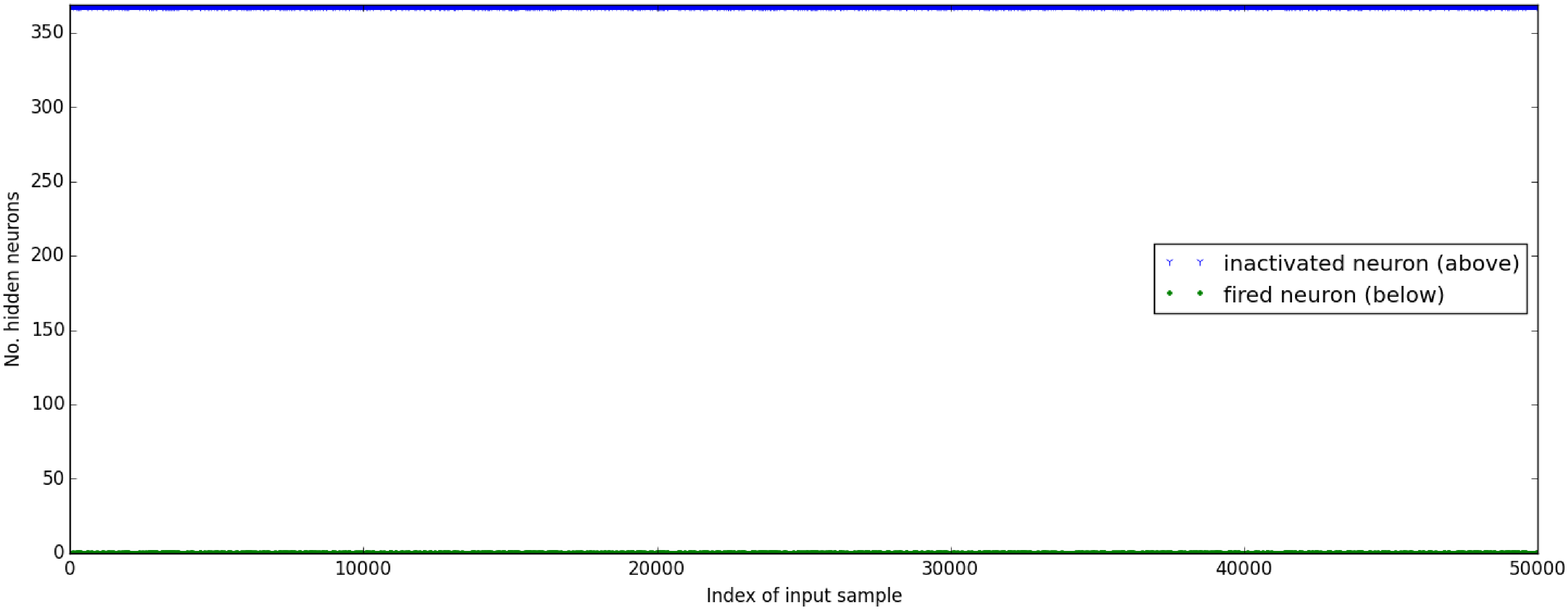}
  \label{fig:dist_h_fired}
}
\subfigure[Inactivated neurons]{\includegraphics[scale=0.22]{dist_h.eps}
  \label{fig:dist_h_inactivated}
}
\vspace{-3mm}
\caption{Distribution of fired or inactivated hidden neurons in layer 3}
\label{fig:dist_h}
\end{figure}

From the discovery described in the section \ref{sec:Knowledge}, the condition of a fired hidden neuron is defined as follows. 
\begin{equation}
  IF \ p(h_j = 1 | \bvec{v}_n) > \theta_{fire}, \ THEN \ neuron \ h_{j} \ is \ fired,\\    
\label{eq:fire_condition}
\end{equation}
where $v_i \in \{0,1\}$, $W_{ij} \in [0, 1]$, $p(h_j = 1 | \bvec{v}_n) \in [0, 1]$ is a conditional probability of $h_j$ under a given $\bvec{v}_n$, which is calculated by sigmoid function. $\theta_{fire}$ is a threshold value to fire. In this paper, we set the $\theta_{fire} = 0.6$. Fig.\ref{fig:dist_h_fired} and Fig.\ref{fig:dist_h_inactivated} show the number of fired and inactivated hidden neurons for each input sample in layer 3. The horizontal axis is the numbering of input samples (up to 50,000). The vertical axis is the number of hidden neurons (up to 369) to each input pattern. From the Fig.\ref{fig:dist_h}, most of the neurons were inactivated, while the number of fired neurons for each input sample was four at most.

\begin{figure}[tbp]
\centering
\includegraphics[scale=1.0]{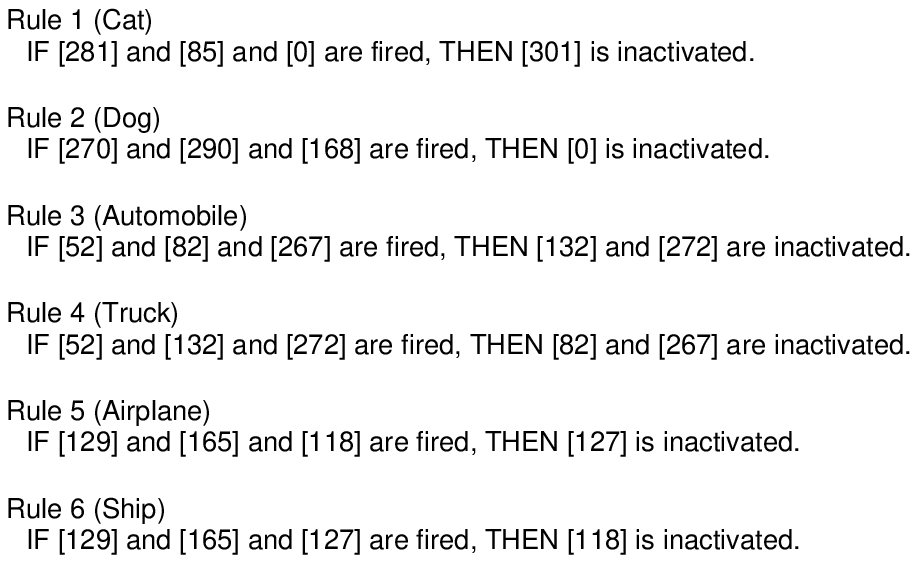}
\vspace{-3mm}
\caption{Acquired Knowledge}
\label{fig:rule}
\end{figure}

The acquired knowledge is 6 kinds rules as shown in Fig.\ref{fig:rule}. The numerical value in `[]' indicates an index of the hidden neuron in layer 3. If the antecedent part at each IF-THEN rule is fired simultaneously for an input pattern, then the signal path to the neuron of the corresponding consequent part will be inactivated. For example, the neurons `281', `85', and `0' in layer 3 are fired at Rule 1, then the signal path to the neuron `301' in layer 4 will be inactivated. If an input pattern is given to the trained network and matches the corresponding neurons in the rules, then the system adapts the IF-THEN influence instead of forward calculation. We examined the classification capability by embedding the rules into the trained network which has 92.4\% correct ratio not 97.1\% in order to confirm the effect of our proposed method. Before embedding the rules, some ambiguous patterns which cause to fire both of antecedent and consequent neurons in the rules were not classified correctly. On the other hand, the calculation result was the highest correct ratio 98.6\% after embedding the rules. As a result, the effectiveness of the proposed synthesis model was shown.

\begin{table}[h]
\caption{Classification Accuracy with acquired knowledge}
\vspace{-5mm}
\label{tab:result-correct-ratio-apply-rule}
\begin{center}
\begin{tabular}{l|r|r}
\hline \hline
& \multicolumn{1}{c|}{Training} & \multicolumn{1}{c}{Test}  \\ \hline\hline
without knowledge&  100.0\%  & 92.4\%   \\ \hline 
with knowledge   &   100.0\%  & {\bf 98.6\%}   \\ \hline 
\hline 
\end{tabular}
\end{center}
\end{table}

\section{Conclusion}
\label{sec:Conclusion}
The adaptive structural learning method of RBM is to self-organize the optimal network structure in terms of energy stability as well as clarification of knowledge according to the given input data during the learning phase. Moreover, we developed assemble method of pre-trained RBM by using the layer generation condition in hierarchical DBN. Our proposed adaptive learning method presents to score a great success for unknown test data on CIFAR-10, but the expensive effort to find the optimal parameter are required in order to classify ambiguous patterns correctly. In this paper, we proposed the fine tuning method that is to switch the signal flow based on knowledge acquired from the trained DBN. 6 kinds of rules were extracted as knowledge and then our proposed method with embedded rules showed the highest classification capability. The fine tuning method will be applied to other datasets such as CIFAR-100 in future work.

\section*{Acknowledgment}
This research and development work was supported by the MIC/SCOPE \#162308002.

\end{document}